\pdfoutput=1

\documentclass[11pt]{article}

\usepackage[]{acl}

\usepackage{times}
\usepackage{latexsym}

\usepackage[T1]{fontenc}

\usepackage[utf8]{inputenc}

\usepackage{microtype}

\usepackage{booktabs}
\usepackage{graphicx}
\usepackage{array}
\newcolumntype{P}[1]{>{\centering\arraybackslash}p{#1}}
\newcolumntype{M}[1]{>{\centering\arraybackslash}m{#1}}
\usepackage{stfloats}
\title{\textit{HADES}: Homologous Automated Document Exploration and~Summarization}

\author{Piotr Wilczyński \and Artur Żółkowski \and Mateusz Krzyziński \\ Warsaw University of Technology  \AND Emilia Wiśnios \and Bartosz Pieliński \and Stanisław Giziński \\ University of Warsaw \\
 \AND Julian Sienkiewicz \and Przemysław Biecek \\ Warsaw University of Technology }

\begin{document}
\maketitle
\begin{abstract}
This paper introduces HADES, a novel tool for automatic comparative documents with similar structures. HADES is designed to streamline the work of professionals dealing with large volumes of documents, such as policy documents, legal acts, and scientific papers. The tool employs a multi-step pipeline that begins with processing PDF documents using topic modeling, summarization, and analysis of the most important words for each topic. The process concludes with an interactive web app with visualizations that facilitate the comparison of the documents. HADES has the potential to significantly improve the productivity of professionals dealing with high volumes of documents, reducing the time and effort required to complete tasks related to comparative document analysis. Our package is publically available on GitHub\footnote{\url{https://github.com/MI2DataLab/HADES}}.
\end{abstract}

\section{Introduction}
The amount of documents produced and stored by individuals, organizations, and governments has been growing exponentially over the past few decades. As the world becomes increasingly digitized, the volume of information generated and processed each day is staggering. This trend has brought many benefits, such as improved access to information and faster dissemination of knowledge. However, it also poses significant challenges for those tasked with managing, analyzing, and understanding this vast quantity of data.

One area where this challenge is particularly acute is in the domain of regulatory compliance. Each country has its own rules and regulations governing various aspects of public life, ranging from environmental protection to labour standards to financial reporting. Compliance with these regulations often involves submitting detailed reports to the relevant authorities, which can be time-consuming and complex.

To address this issue, the European Union recently introduced new regulations mandating the creation of a centralized reporting system for each country\footnote{\url{https://ec.europa.eu/isa2/sites/default/files/eif_brochure_final.pdf} (last accessed: 25.02.2023)}. This system is designed to streamline the reporting process, making it easier for businesses and organizations to comply with regulatory requirements. However, even with this new system in place, the volume of documents that must be managed and analyzed will continue to grow.

\begin{figure*}[t!]
    \centering
    \includegraphics[scale=0.35]{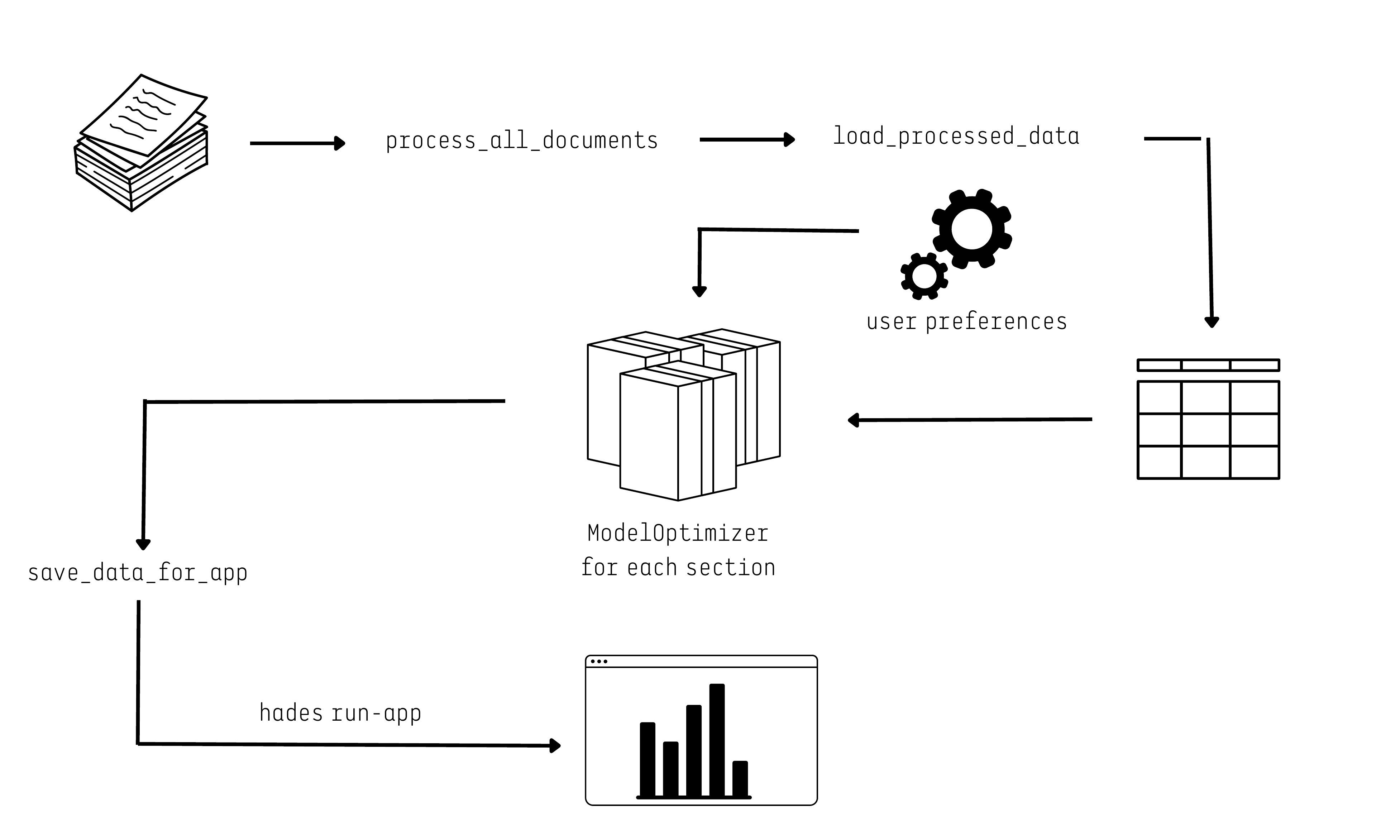}
    \caption{Diagram of HADES pipeline.}
    \label{fig:pipeline}
\end{figure*}

To help address this challenge, we present HADES, a novel tool for automatic comparative documents with similar structures. HADES is designed to streamline the work of professionals dealing with large volumes of documents, such as policy documents, legal acts, and scientific papers. The tool employs a multi-step pipeline that begins with processing PDF documents using topic modeling, summarization, and analysis of the most important words for each topic. The process concludes with an interactive web app with visualizations that facilitate the comparison of the documents. HADES has the potential to significantly improve the productivity of professionals dealing with high volumes of documents, reducing the time and effort required to complete tasks related to comparative document analysis. In the following sections, we describe the design and implementation of HADES in detail and evaluate its performance using the European Union (EU) documents - Analysis of National recovery and resilience plans\footnote{\url{https://commission.europa.eu/business-economy-euro/economic-recovery/recovery-and-resilience-facility_en} (last accessed: 18.02.2023)}, National Energy and Climate Plans\footnote{\url{https://commission.europa.eu/energy-climate-change-environment/implementation-eu-countries/energy-and-climate-governance-and-reporting/national-energy-and-climate-plans_en} (last accessed: 21.02.2023)}.

\section{Related work}
Topic modeling has been widely used in natural language processing to extract meaningful topics from a corpus of text data. However, the resulting models can be complex and challenging to interpret. Visualization has emerged as a powerful tool for exploring and presenting topic modeling results. Various visualization techniques have been proposed to help researchers and practitioners gain insights into the underlying structure of the data. Some approaches are based on the use of well-known types of visualizations, such as wordclouds and barplots. Worth noting are more advanced techniques like Termite \cite{termite}, which provides a term-topic matrix showing term distributions for all latent topics, or LDAvis \cite{ldavis}, which uses interactive visualizations. HADES provides a more holistic approach through the availability of multiple types of visualization simultaneously, both simple and more complex.

Our previous work \cite{climate_policy_tracker} served as a prototype for HADES, which was originally designed to cater specifically to National Energy and Climate Plans. However, the pipeline used in the prototype lacked some key features, such as a summarization module and limited topic modeling methods, and the frontend of the application was rudimentary. As a result, HADES has been developed as a more comprehensive and versatile tool, capable of analyzing various structured documents beyond National Energy and Climate Plans.

\section{Architecture and design}
A simplified diagram of the subsequent steps of the proposed pipeline is shown in Figure \ref{fig:pipeline}. It contains the successive commands required to create an application to summarize the results. A detailed description of the main HADES modules is presented below.

\subsection{Preprocessing Module}
HADES provides semi-automatic preparation of documents for further analysis. Text is extracted from PDF files with \texttt{pypdf2} Python package\footnote{\url{https://github.com/py-pdf/pypdf}} based on the table of contents or defined metadata about the structure of the documents provided manually. Creating a corpus is also possible with this step omitted when the texts are already extracted. Further functionality allows preprocessing of texts into the form used in topic modeling (i.e., removal of stopwords, tokenization, lemmatization). This process also can be replaced by the manual generation of the corpus in the form of a well-defined data schema.

\begin{table*}[!b]
    \centering
    \begin{tabular}{M{0.15\textwidth}M{0.2\textwidth} c m{0.4\textwidth}}
    \toprule
    \textbf{Area} & \textbf{Visualization} & 
    & \textbf{Description} \\ \midrule
        Clustering & Geovisualization  & 
        \begin{minipage}{.12\textwidth}
          \includegraphics[width=\linewidth]{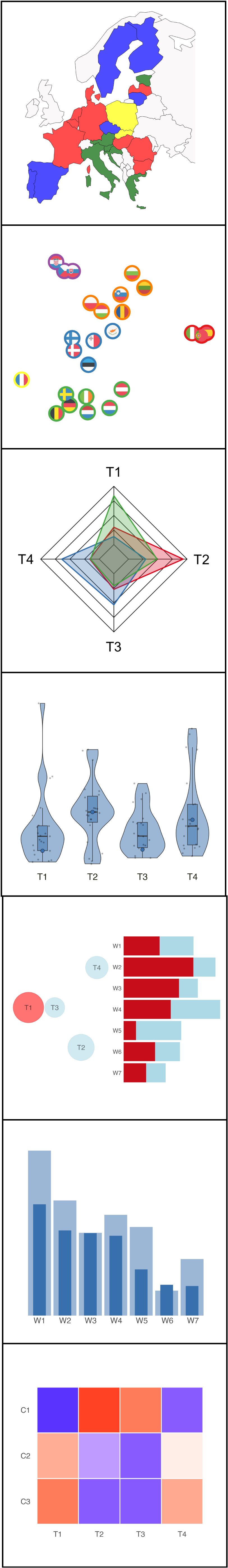}
        \end{minipage}
        &  In the case where the entities creating the documents are different countries/states, it allows for examining the relationship between geopolitics and the clustering based on the obtained modeling. \\
    \hline 
        Clustering & Dimensionality reduction mapping & 
        \begin{minipage}{.12\textwidth}
          \includegraphics[width=\linewidth]{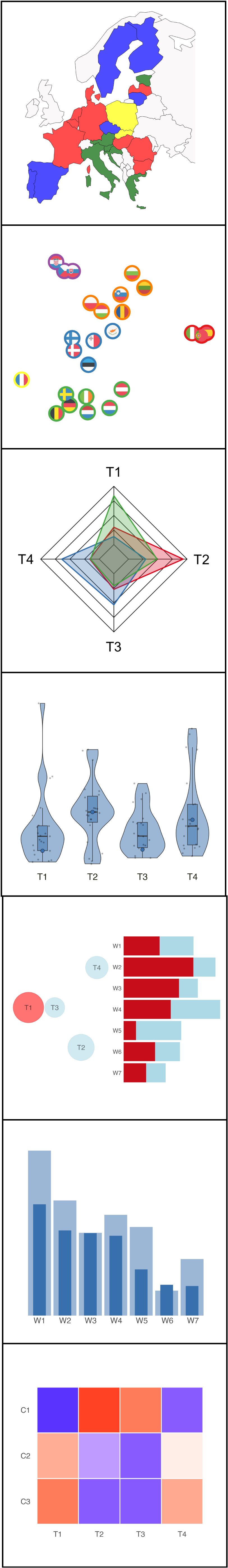}
        \end{minipage}
        & It enables a more universal visualization of the clustering and preserves the meaning of the distance between entities.  \\
    \hline 
        Inter-document analysis & Radar plot of topic distributions for selected documents & 
        \begin{minipage}{.12\textwidth}
          \includegraphics[width=\linewidth]{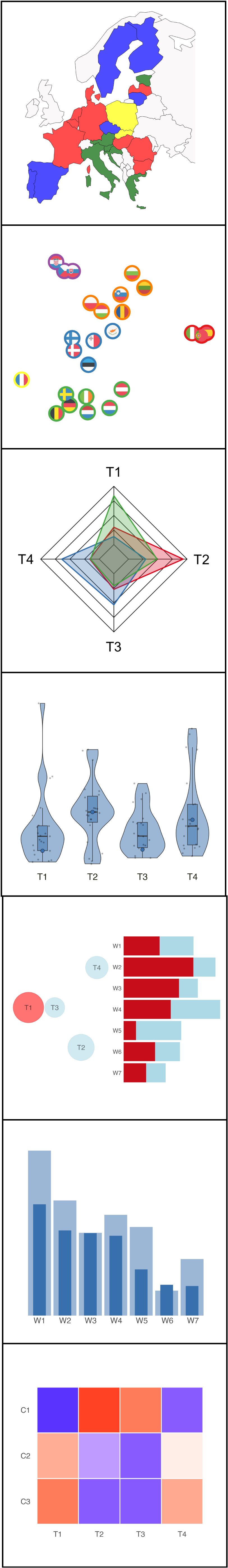}
        \end{minipage}
        &  It allows for a detailed comparison of the obtained distribution of topics for specific entities. \\
    \hline 
        Inter-document analysis & Violin plot of topic distribution across all documents & 
        \begin{minipage}{.12\textwidth}
          \includegraphics[width=\linewidth]{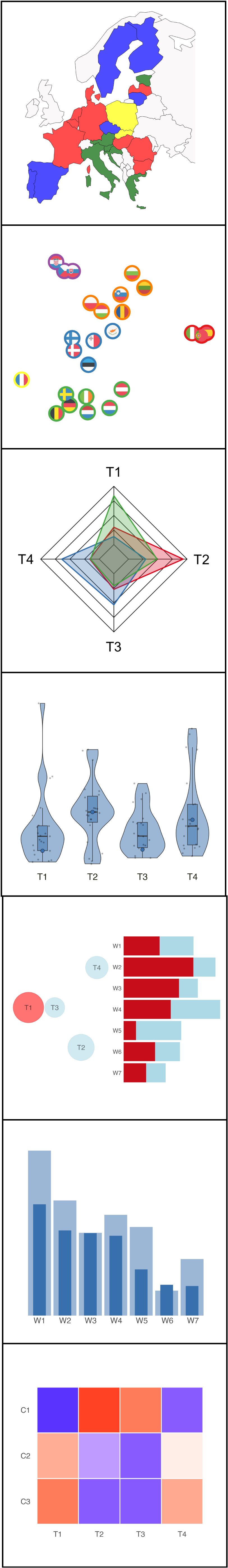}
        \end{minipage}
        &  It allows relating the obtained distribution of topics (e.g., political agenda) to the entire analyzed set. \\
    \hline 
        Intra-topic analysis & LDAvis & 
        \begin{minipage}{.12\textwidth}
          \includegraphics[width=\linewidth]{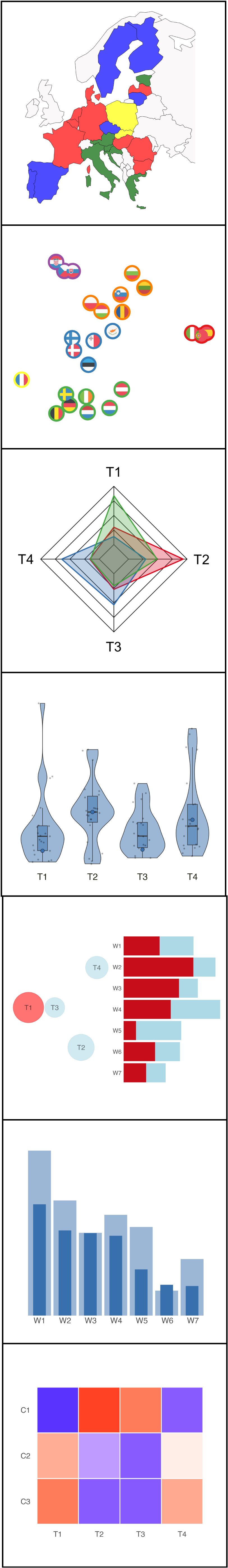}
        \end{minipage}
        & It enables interactive visualization and exploration of~topics. \\
    \hline 
        Intra-topic analysis & Bar plot of keywords for selected topics & 
        \begin{minipage}{.12\textwidth}
          \includegraphics[width=\linewidth]{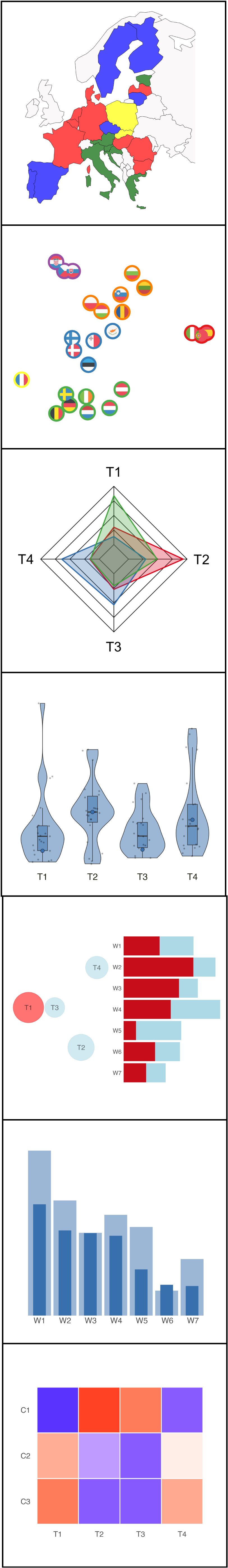}
        \end{minipage}
        &  It allows for a thorough investigation of topics based on keywords. \\
    \hline 
        Correlation analysis & Correlation matrix & 
        \begin{minipage}{.12\textwidth}
          \includegraphics[width=\linewidth]{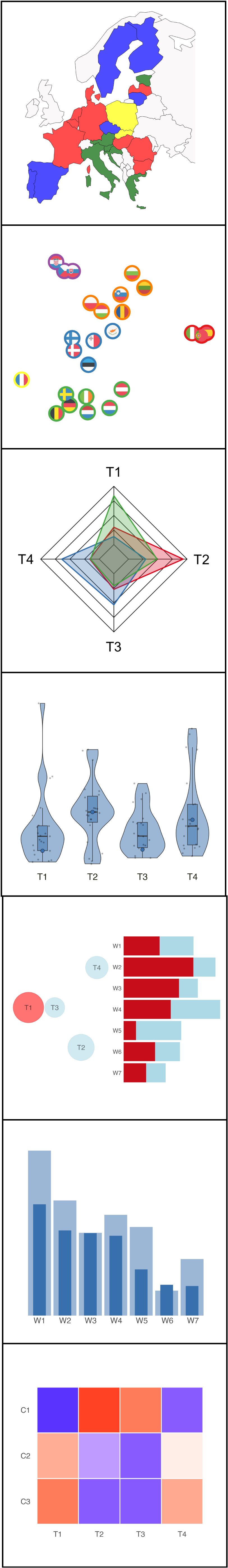}
        \end{minipage} & It allows for examination of the dependency between the obtained topics probabilities and other covariates.
    \\
    \bottomrule
    \end{tabular}
    \caption{A summary of the visualizations available in HADES to explore the obtained topic modeling.}
    \label{tab:vis}
\end{table*}

\subsection{Topic Modeling Module}
Our pipeline offers three distinct topic modeling approaches with the option to add additional models. The first approach is Latent Dirichlet Allocation (LDA) \cite{lda} from the \texttt{Genism} package \cite{genism}. This widely used method assumes that each document in a corpus is a combination of multiple topics, where each topic is a probability distribution over words. LDA works by randomly assigning each word in each document to a topic and then refining those assignments iteratively by determining the probability that each word belongs to each topic and adjusting the topic assignments accordingly. This process continues until the algorithm converges on a set of topic assignments that best explain the distribution of words across the documents in the corpus.

The second approach is the Non-Negative Matrix Factorization (NMF) \cite{nmf} from the \texttt{Genism} package \cite{genism}. It involves decomposing a large matrix of word frequencies into a smaller matrix that captures the underlying topics of the text. NMF factorizes the original matrix into two non-negative matrices: a term-topic matrix and a topic-document matrix. The term-topic matrix represents the relative importance of each term in each topic, while the topic-document matrix represents the degree to which each document is related to each topic. The NMF algorithm updates the values in the two matrices iteratively to minimize the error between the original matrix and the low-rank approximation, subject to the constraint that all values in the matrices must be non-negative.

The final model available is Contextualized Topic Modeling \cite{ctm}, which incorporates contextualized representations in topic models and is comprised of two main components: (i) the neural topic model ProdLDA \cite{prodlda} based on the Variational AutoEncoder (VAE) and (ii) the SBERT \cite{sentencebert} embedded representations. The model is independent of the choice of topic model and pre-trained representations as long as the topic model extends an autoencoder and the pre-trained representations embed the documents. The contextualized document embeddings from SBERT are concatenated with the BoW representation, projected through a hidden layer with the same dimensionality as the vocabulary size, and extended in ProdLDA to include a product of experts instead of the multinomial distribution over individual words in LDA.

\subsection{Analysis Module}
The proposed pipeline includes a number of functionalities to analyze the obtained topic modeling. One of them is clustering based on distributions of topics for individual entities. The clustering algorithms used are hierarchical clustering \cite{johnson1967hierarchical}, k-means \cite{lloyd1982least-kmeans} and HDBSCAN \cite{hdbscan}. Jensen–Shannon divergence and Hellinger distance, originally described for LDA model \cite{similarity-measures}, can be used as distance metrics. The significance of differences between topic distributions in clusters is tested using MANOVA. 

Further analysis of the obtained modeling is possible through visualizations adapted to the comparative analysis of topic distributions between documents. Interpretation of the topics themselves is made possible through the use of the LDAvis tool \cite{ldavis}. This interactive method created directly for the LDA model but with the possibility of using it for other models, provides a global overview and the variation of the topics and illustrates how they differ. At the same time, it allows for a scrutinous examination of the terms most associated with each topic based on relevance and saliency metrics \cite{saliency}.

Pipeline also allows examining the correlation between the probability of a given topic and other covariates, which may be metadata about entities that are associated with the documents. Breakdown and descriptions of visualizations available in HADES are shown in Table \ref{tab:vis}.

Interpretation of the documents is further facilitated by prepared summaries, the process of generating which is described in Section \ref{subsec:summarization}.

All analytical functionalities are available in the automatically created document analysis application, see Section \ref{subsec:app}.  

\subsection{Summarization Module}
\label{subsec:summarization}
Our summarization module adopts a hybrid approach that combines both extractive and abstractive summarization techniques. Firstly, we utilize BERT extractive summarization \cite{bert_extractive_summary} to identify the most significant sentences from the original text. This is necessary due to two reasons: firstly, the input size for the GPT-3 model \cite{gpt3} used in the abstractive part is limited, and secondly, previous studies have demonstrated that the length of input can influence the degree of hallucination in pretrained language models \cite{gpt3_hallucinations}. The selected sentences are then utilized to produce an abstractive summary with the prompt \texttt{Summarize the text above in three sentences}. We opted to limit the number of sentences in the prompt as it produced the most favourable experimental outcomes. We also evaluated other prompts such as \texttt{tl;dr} and \texttt{summarize the text above} but found them to be unsatisfactory.

\subsection{Application}
\label{subsec:app}
\begin{figure}[h]
    \centering
    \includegraphics[width=0.45\textwidth]{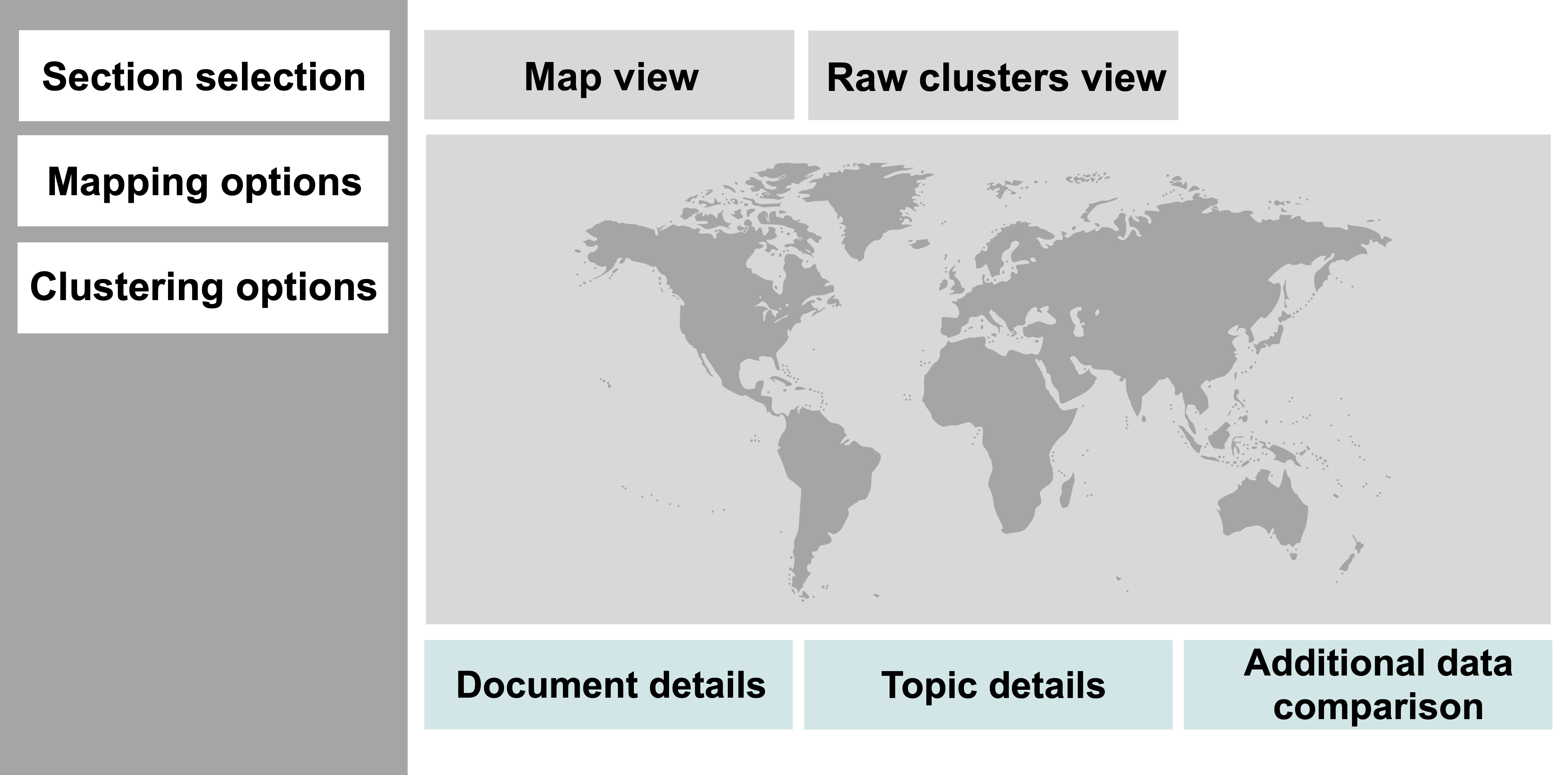}
    \caption{Schematic view of HADES application.}
    \label{fig:app}
\end{figure}

Our web-based solution, developed using \texttt{Streamlit}\footnote{\url{https://streamlit.io/}}, consists of four primary interactive sections (see Figure \ref{fig:app} for schematic view). The left-side panel allows users to select document sections and customize mapping (tSNE \cite{tsne}, UMAP \cite{umap}), clustering (hierarchical \cite{johnson1967hierarchical}, k-means \cite{lloyd1982least-kmeans}, HDBSCAN \cite{hdbscan}), and their parameters. 

The central section of the application displays the primary plot representing clustering. If the document IDs correspond to countries, a map plot is available. Otherwise, only a 2D raw cluster view can be accessed. 

Below the plot, users can access three detailed views: document details, topic details, and additional data comparison. The document details view provides a summary for each country. To enhance the interpretability of our topic modeling methods, we present the most important sentences and words for each topic. We display several sentences with marked probabilities of the most important words for each topic in each section. Additionally, users can compare multiple document sections. After selecting the document IDs from the list, the radar plot shows the topic distribution for each document. Meanwhile, on the right side of the screen, the violin plot displays the distribution of each topic. The violin plots also indicate the position of each document in the distribution compared to the others.

\section{System usage}
The pipeline generates all the necessary data to initiate the application. The \texttt{process\_all\_documents} function is accountable for extracting text from pdf files and storing them in txt format. These files are then passed through the \texttt{load\_processed\_data} function. If they are being loaded for the first time, they are preprocessed and stored accordingly.

Subsequently, the \texttt{model\_optimizer} class objects, which are the core components of the pipeline, can be constructed with the prepared data. These objects should correspond to specific sections or labeled segments of text in the documents. During initialization, the user can specify the number of topics required, and select the desired model and its parameters. The \texttt{model\_optimizer} identifies the optimal model that fulfils the user's criteria. Once the object is generated, the user can assign labels to the identified topics either manually or using the GPT3 algorithm \cite{gpt3}. The \texttt{model\_optimizer} class provides various information regarding the generated model and the distribution of topics.

After adjusting the \texttt{model\_optimizers}, invoking the \texttt{save\_data\_for\_app} function will store the processed data obtained from the generated models in a format compatible with the application. Finally, running the command \texttt{hades run-app --config config.json} in the terminal is sufficient to launch the application.

\section{Case Study}
In order to demonstrate the efficacy of our proposed approach, we opted to evaluate it on two distinct document collections, namely, National Energy and Climate Plans, and Analysis of Recovery and Resilience Plans. We intentionally chose the former dataset as it had been previously studied in \cite{climate_policy_tracker} and would enable us to showcase enhancements achieved by our solution. The latter dataset, on the other hand, would demonstrate the generalizability of our method. To enable comparison, we selected Austria as the target country for analysis in both datasets.

\subsection{National
Energy and Climate Plans (NECP)\footnote{The demo app for NECP is available here: \url{https://krzyzinskim-hades-hades-appmain-dxm7g7.streamlit.app/}}}

NECPs provide details on the approaches adopted by EU Member States pertaining to the critical energy and climate policy dimensions. These documents were mandated to adhere to a prescribed structure, which necessitated their categorization into five sections and five dimensions. It was essential to process these documents to demonstrate that our pipeline could analyze not only individual sections but also other labeled components of the documents.

The summary for Austria, depicted in Appendix \ref{appendix:au_summary}, highlights the objective of achieving energy production exclusively from renewable sources by the year 2030. In the EU-wide evaluation of National Energy and Climate Plans\footnote{\url{https://eur-lex.europa.eu/legal-content/EN/TXT/PDF/?uri=CELEX:52020DC0564&from=EN} (last accessed: 25.02.2023)}, which provides a comprehensive overview of individual assessments, Austria is referenced only four times, predominantly in this specific objective's context. 

The EU-wide assessment report states that "a few Member States have established highly ambitious sectoral targets for renewables, with Austria and Sweden aiming for 100\% renewable electricity by 2030 and 2040, respectively." In our comparative analysis of Austria and Sweden, we observed that the two countries shared similar views on regulatory procedures and climate action plans (see Appendix \ref{appendix:plots}).

Additionally, Austria is mentioned in the context of energy poverty in the same EU-wide assessment report. The report highlights that "several Member States use primary indicators developed by the European Energy Poverty Observatory, with NECPs addressing affordability issues in the context of energy and climate transitions. Countries like Austria, Belgium, France, the Netherlands, and Denmark have all included measures to address this issue." In our comparative analysis of the Energy Security section, we found that these countries share many similarities in terms of their policies and approaches.

\subsection{Analysis of Recovery and Resilience Plans (ARRP)}
The National Recovery and Resilience Plan aims to revive the economy in the aftermath of the pandemic by implementing reforms that promote sustainability and adaptability. The plan focuses on six core pillars, including green transition, digital transformation, and economic competitiveness. 

The ARRP summary report\footnote{\url{https://commission.europa.eu/system/files/2022-07/com_2022_383_1_en.pdf} (last access: 25.02.2023)} is not amenable to the type of analysis afforded by the NCEP. As a result, our evaluation of the system will focus solely on the available summaries. The appendix \ref{appendix:arrp} contain summaries for all sections concerning Austria. According to the joint report, the primary outcome of Austria's recovery plan following the crisis is the reduction of the gender pay gap and the improvement of working conditions for professions predominantly occupied by women. This conclusion is in agreement with the summaries provided by HADES, particularly in the section concerning general conclusions from the document.

\section{Limitations and Future Directions}
\paragraph{Document Structure} Our present approach is currently optimized for well-structured documents, enabling the comparison of multiple documents on the same topic. Moving forward, we aim to broaden the applicability of our approach to all types of documents by identifying pertinent textual fragments.
\paragraph{PDF Processing} The PyPDF2 library, used for automated document parsing, generates numerous artefacts in the text, particularly when processing tables, image captions, and other elements. In future work, we intend to adopt a more sophisticated approach incorporating Document Layout Analysis to isolate only the textual content.

\section{Conclusions}
This paper presents HADES, an innovative tool that employs a multi-step pipeline to automatically analyze and compare documents with similar structures. HADES has the potential to significantly enhance the productivity of professionals dealing with high volumes of documents, such as policy-makers, legal experts, and scientists. By leveraging topic modeling, summarization, and analysis of the most important words for each topic, HADES streamlines the processing of large amounts of textual data and generates an interactive web app with visualizations that facilitate the comparison of the documents. This open-source solution can help professionals make quick decisions on what to focus on in their work, without the need for manual checking of every document for similarities and differences. Overall, HADES represents a significant advancement in comparative document analysis, and its public availability on GitHub makes it a valuable resource for professionals dealing with large volumes of documents.

\section*{Acknowledgements}
Research was funded by (POB Cybersecurity and Data Science) of Warsaw University of Technology within the Excellence Initiative: Research University (IDUB) programme.

\section*{Ethical Impact Statements}

The purpose of this ethical impact statement is to outline the potential ethical implications of the use of HADES. The purpose of this tool is to facilitate the analysis of multiple documents with a similar structure, such as public policies or legal contracts. The tool uses natural language processing and machine learning algorithms to compare and identify document similarities and differences.

The development and use of this tool raise several ethical considerations. The following is a non-exhaustive list of potential ethical issues:
\par 
\textbf{Data Privacy} The tool uses natural language processing algorithms to compare documents and therefore requires access to the text of the documents being analyzed. This raises concerns about data privacy, particularly if the documents contain sensitive or personal information. Developers must ensure that appropriate measures are in place to protect the privacy and confidentiality of the documents and the data they contain.
\par 
\textbf{Bias and Fairness} Machine learning algorithms are only as unbiased as the data they are trained on. It is possible that the tool may produce biased results if it is trained on a limited or skewed dataset. Developers must ensure that the training data is diverse and representative of the population of documents the tool intends to analyze.

\bibliography{bibliography}
\bibliographystyle{acl_natbib}

\appendix

\section{Austria's summary for NECP for decarbonization section}
\label{appendix:au_summary}

The Austrian government has set a goal to get 100\% of its electricity from renewable sources by 2030. They are planning on doing this by increasing energy efficiency in the agricultural sector, electrifying applications near farms, and by using damaged wood to produce Fischer-Tropsch diesel.

\section{Additional plots from the application}
\label{appendix:plots}

\begin{figure}[h]
    \centering
    \includegraphics[width=0.45\textwidth]{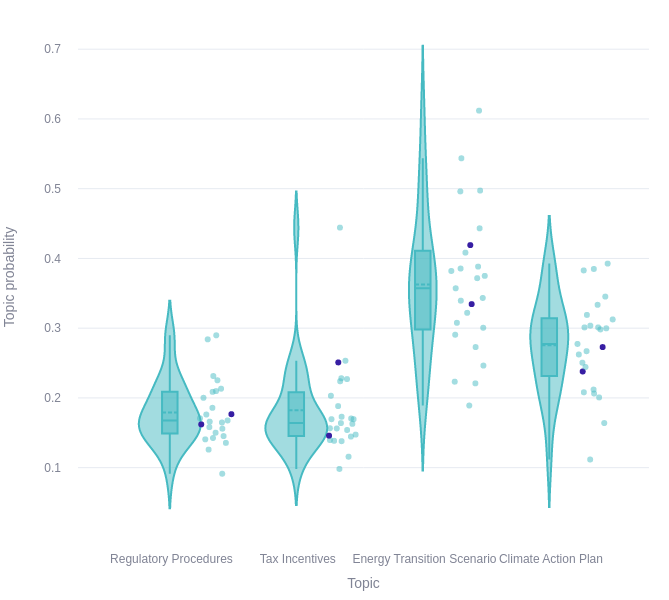}
    \caption{Plot from the application showing a comparison between Austria and Sweden in the decarbonization section.}
    \label{fig:au_sw}
\end{figure}

\begin{figure}[h]
    \centering
    \includegraphics[width=0.45\textwidth]{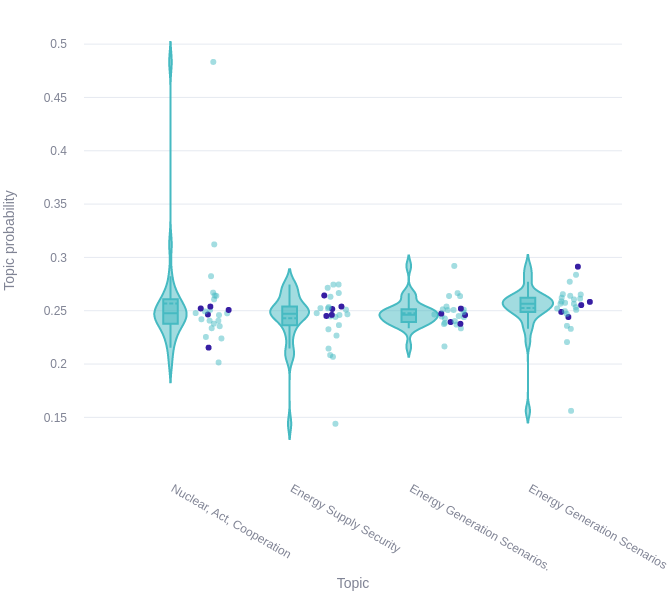}
    \caption{Plot from the application showing a comparison between Austria, Belgium, France, the Netherlands, and Denmark in the energy security section.}
    \label{fig:energy_poverty}
\end{figure}

\section{Austria's summaries for ARRP}
\label{appendix:arrp}

\noindent
\textit{Executive summary:}
\par \noindent
Austria's economy was negatively impacted by the COVID-19 crisis, causing the unemployment rate to increase to 5.4\%. The country has implemented a recovery and resilience plan with sound underlying assumptions to ensure environmental objectives are met. The Ministry of Finance is the central coordinating body for this system.
\par \noindent
\textit{Recovery and resilience challenges: scene-setter:}
\par \noindent
The Austrian recovery and resilience plan focuses on inclusive growth and has two reform elements in the subcomponent of Arts and Culture. It makes explicit reference to the consistency with other programmes, and does not include a security self-assessment for investments in connectivity and digital capacities.
\par \noindent
\textit{Objectives, structure and governance of the plan:}
\par \noindent
Austria's economy was strongly affected by the COVID-19 pandemic in 2020, with public spending increasing to mitigate its socio-economic consequences. Austria has considerable scope to green its tax system and take targeted measures in digital public services to deal with the crisis. The country needs to make more efforts to achieve carbon neutrality and increase its performance in the digital transition.
\par \noindent
\textit{Summary of the assessment of the plan:}
\par \noindent
The Austrian recovery and resilience plan provides for a comprehensive set of measures that effectively addresses the six pillars referred to in the RRF Regulation. This plan will positively contribute to the growth potential and enhance the competitiveness of the Austrian economy, and reduce social vulnerabilities. It will also contribute to the implementation of the European Pillar of Social Rights and improve the indicators of the Social Scoreboard.







\end{document}